\documentclass[11pt,a4paper]{article}
\usepackage[hyperref]{main}
\usepackage{times}
\usepackage{latexsym}
\usepackage{tabularx}
\usepackage{enumitem}
\usepackage{graphicx}
\usepackage{multicol}
\usepackage{textpos}
\usepackage{booktabs}

\usepackage{url}

\aclfinalcopy

\title{Towards better substitution-based word sense induction}

\author{Asaf Amrami$^{~\dagger}$ \and Yoav Goldberg$^{~\dagger~\ddagger}$ \\[0.5em]
  $\dagger$ Computer Science Department, Bar Ilan University, Israel \\
  $\ddagger$ Allen Institute for Artificial Intelligence \\[0.5em]
  {\tt \{asaf.amrami, yoav.goldberg\}@gmail.com}
  }

\date{}

\begin{document}
\maketitle
\begin{abstract}
 Word sense induction (WSI) is the task of unsupervised clustering of word usages within a sentence to distinguish senses. Recent work obtain strong results by clustering lexical substitutes derived from pre-trained RNN language models (ELMo). Adapting the method to BERT improves the scores even further. We extend the previous method to support a dynamic rather than a fixed number of clusters as supported by other prominent methods, and propose a method for interpreting the resulting clusters by associating them with their most informative substitutes. We then perform extensive error analysis revealing the remaining sources of errors in the WSI task.
 Our code is available at \url{https://github.com/asafamr/bertwsi}.
\end{abstract}

\section{Introduction}

\paragraph{Word Sense Induction }

Word Sense Induction (WSI) is the task of clustering in-context usages of
words to groups that represent senses.
A WSI system is given multiple sentences containing usages (instances) of
target lemma+part-of-speech, and is expected to group together usages in which
the target is used in the same sense. E.g., in:

\begin{enumerate}[itemsep=-2mm]
\item I like \textbf{warm} summer evenings
\item They were greeted by a \textbf{warm} welcome
\item The waters of the lake are \textbf{warm}
\end{enumerate}
we would like to group (1) and (3) into one sense and (2) to a different
one.\footnote{Things are often more challenging: a \textit{\textbf{warm} hand},
for example, might belong to either of the senses above, requiring larger
context to disambiguate.}

WSI was explored in several SemEval shared tasks
\cite{semeval2007wsi,semeval2010wsi,semeval2013wsi}, with the gold-labels
following human annotation according to the WordNet \cite{wordnet} sense inventory.
WordNet senses are very fine-grained and often hard to tag even for experts \citep{snyder2004english}, leading the latest evaluation (SemEval 2013 task 13) to replace
the hard-clustering task with a soft-clustering one, in which each instance can simultaneously belong to several clusters, each with a different label. This clustering is then compared to
the human-taggers' disagreement data.

Up until recently, state-of-the-art results for WSI were dominated by a series
of increasingly sophisticated graphical models
\cite{lau2013unimelb,wang2015sense,komninos2016structured,hwang2019autosense}. 

A competing approach relies on \emph{substitute vectors}: each target instance
is represented by a distribution over possible in-context probable substitutes for the
word, and clustering is performed over these distributions. This approach by
\citet{baskaya2013ai} was implemented in the AI-KU system using n-gram language
models (LM).

In recent work 
\cite{amrami2018word}, henceforth referred to as LSDP (Language-model Substitution with Dynamic Patterns), we showed that
by replacing the n-gram LM with ELMo-based biLM \cite{peters2018deep} and adding a \emph{dynamic
patterns} technique, the substitute vectors approach can achieve
state-of-the-art results (Section \ref{sec:elmosp}).

In this work, we further explore the use of substitute-based approaches for WSI. After verifying that the approach transfers to the recently introduced \textsc{Bert} deep masked LM \cite{devlin2018bert} (with a very significant improvement in WSI scores), we make two additional contributions to the mentioned method: (a) we present a method to move from a fixed number of senses across target words to choosing a dynamic number of senses for each target (as supported by most other WSI methods, e.g., \cite{teh2005sharing,komninos2016structured,hwang2019autosense}; and (b) we suggest to use the substitutes as a mean of interpreting / analyzing the resulting sense cluster by considering prominent word substitutes. This enables a more in-depth error analysis, showcasing and quantifying the remaining kinds of errors in WSI.

\section{The LSDP WSI Method}

\label{sec:elmosp}
We describe the LSDP-algorithm, which we extend.
Given $k$ in-sentence instances of a target word which we wish to cluster into senses, each instance is associated with $r$
representatives. Each representative is composed of $n$ words, which are
sampled with replacement from the LM.\footnote{Sampling is performed from the
softmax distribution over the top-$\ell$ logits, and while ignoring the bias
terms.} The sampled words are lemmatized, and each representative is then represented as
a one-hot vector of its lemmas (multiple
occurrences of the same lemma within a representative are discarded). The
resulting set of $k*r$ 1-hot vectors goes through TFIDF transform to
discount uninformative words, and the resulting vectors are clustered into a
predefined number of clusters using
hierarchical clustering with cosine distance and average linkage. This provides a \emph{hard clustering} over \emph{representatives}. This clustering is converted to a 
\emph{soft-clustering} over \emph{instances}, by associating each instance to a cluster according to the
percentage of its representative that are assigned to that cluster.

Sampling the representatives from the ELMo biLM does not take into account the
word itself when predicting substitutes. The \textbf{dynamic patterns} approach
in LSDP overcomes this by querying the LM for a linguistically motivated
manipulated context that
take the target word into account. 
By way of example, to get substitutes for \emph{brown} in \emph{``my dogs are
brown''}, the forward LM is presented with \emph{``my dogs are \underline{brown and} $\Box$''} rather than with \emph{``my dogs are $\Box$''}. This encourages relevant substitutes
such as \emph{black} and discourages less desirable ones like \emph{barking}.


\section{Better Substitution Based WSI}

\subsection{From ELMo to Bert}
Contextualized vector representations from the recently introduced \textsc{Bert} model \cite{devlin2018bert} were shown to outperform ELMo on several NLP tasks. Like ELMo, \textsc{Bert} is trained in a self-supervised manner to predict words in a sentence. \textsc{Bert} differs from ELMo by being based on a Transformer \cite{vaswani2017attention} instead of an LSTM, being truly bidirectional deep model, and, most importantly for our purposes, being trained in a ``noisy masked LM'' setup in which the training procedure receives a sentence, replaces some words with a MASK symbol and randomly perturbs some others, and then attempts to predict the original words from the resulting text. Thus, the model learns representations which are predictive of the words in context, and can also take into account the current word when making a prediction. For further details, see \cite{devlin2018bert}.

The dynamic patterns in LSDP were motivated by the BiLM not ``seeing'' the
target word otherwise. In contrast, \textsc{Bert}'s architecture and training procedure
does allow providing the word together with its context when predicting
substitutes. 
Simply replacing the ELMo LM with BERT's, without dynamic patterns, already provides state-of-the-art
results for the SemEval 2013 task, with an AVG score of 37.0 compared to 25.4 with
the full LSDP.
However, during BERT training words are sometimes randomly replaced, making
BERT suspicious and often predicting substitutes by context alone: a probable
substitute for \emph{``my [dogs] are brown''} according to BERT is \emph{eyes}, which is clearly not a lexical substitute for dog.

This suggests BERT could also benefit from \textbf{dynamic patterns}. However,
the pattern from \citet{amrami2018word} did not show big improvements as-is. 
Instead, we use parenthetical
patterns. 
We empirically found that the pattern ``\emph{target ( or even [MASK] )}'' (e.g.
\emph{my dogs (or even [MASK]) are brown}) yields
good results.\footnote{Among others, our search for other patterns included parenthesized and apposition based lexical substitute promoting patterns such as \emph{and [MASK]},\emph{or [MASK]} as well as hyponymy promoting patterns like \emph{such as [MASK]}} 

This pattern resulted in similar scores to using the
vanilla sentence and predicting over the target token, while yielding somewhat
different results. We combine the two by averaging the logits from both
predictions prior to the softmax.\footnote{We also further divide the inputs
logits by a temperature parameter to smooth the distribution and diversify the
words. We also lemmatize BERT word pieces to predicted lemmatized words, and pad
the sentence with the [CLS] and [SEP] tokens as required by BERT.}
\\
\noindent\textbf{Results }
Evaluation on SemEval 2013 and SemEval 2010 yield 
state-of-the-art results on both datasets mainly due to BERT powerful LM (Tables \ref{tab:results},\ref{tab:results2010}). The addition of dynamic patterns contribute almost 2 additional points to the already
high AVG scores.\footnote{To comply with the hard-clustering setup of SemEval
2010, we compute the soft-clustering and take the most probable sense for each
instance. The hard clustering also suggests the use of V-measure and F-score
instead of FNMI and FBC for SemEval 2010. We use $\ell=200$, $temperature=1.25$,
$nReps=15$, $nSenses=7$, $minInstances=2$, $nRepSamps=20$. The compared systems
are: AutoSense \cite{hwang2019autosense}, 
MCC-S \cite{komninos2016structured}, 
SenseTopic \cite{wang2015sense}, 
SE-WSI-fix \cite{song2016word}.
AI-KU \cite{baskaya2013ai}
, BNP-HC \cite{teh2005sharing}
, LDA \cite{blei2003latent}. Numbers taken from the corresponding publications.}

\begin{table}[t!]
\centering
\scalebox{0.8}{
\begin{tabular}{| l | l | l | l |}
	\hline
  Model & FNMI & FBC & AVG \\
  \hline 
  {\bf Ours} &{\bf 21.4 \small (0.5)} &{\bf 64.0 \small  (0.5) } &{\bf 37.0 \small (0.5)} \\ 
  {\bf Ours:ND} & 19.3  \small (0.7) & 63.6 \small (0.2) & 35.1 \small (0.6)\\
  LSDP & 11.3  & 57.5 & 25.4  \\
  AutoSense & 7.96 &  61.7 & 22.16 \\
  MCC-S & 7.62 & 55.6 & 20.58 \\
  ST(SW)& 7.14 & 55.4 & 19.89 \\
  AI-KU  & 6.5 & 39.0 & 15.92 \\
  \hline 
\end{tabular}}
\caption{Evaluation Results on the SemEval 2013 Task 13 Dataset. We report our mean (STD) scores over 10 runs. \textbf{ND}: no dynamic patterns. \textbf{ST(SW)}: Sense-Topic with embedding similarity weighting.
}
  \label{tab:results}
\end{table}

\begin{table}[t!]
\centering
\scalebox{0.8}{
\begin{tabular}{| l | l | l | l |}
	\hline
  Model & F-S & V-M & AVG \\
  \hline

  {\bf Ours} &{\bf  71.3 \small (0.1) } &{\bf 40.4 \small (1.8) } &{\bf 53.6 \small (1.2)}\\
  {\bf Ours:ND} & 70.9 \small (0.4) & 37.8 \small (1.5) & 51.7 \small (1.2) \\
  AutoSense & 61.7 &  9.8 & 24.59 \\
  SE-WSI-fix &  55.1 & 9.8 & 23.24\\
  BNP-HC & 23.1 & 21.4 & 22.23 \\
  LDA  & 60.7 & 4.4 & 16.34 \\
  \hline
\end{tabular}}

\caption{Evaluation Results on the SemEval 2010 Task 14 Dataset. We report our mean (STD) scores over 10 runs. \textbf{ND}: no dynamic patterns.}
  \label{tab:results2010}
\end{table}

\subsection{Dynamic Number of Clusters}
\label{sec:numc}
LSDP uses a fixed number of 7 clusters for all target words, a choice which was shown to work well on the SemEval 2013 task 13 dataset. However, using a fixed number this way is obviously sub-optimal.\footnote{Indeed, other works on WSI \cite{teh2005sharing,komninos2016structured,hwang2019autosense} does attempt to infer the number of clusters for each sense. \citet{teh2005sharing} do it by employing a stick breaking clustering process and \citet{komninos2016structured} use a model selection criteria to prevent over specification.\citet{hwang2019autosense} preform a cleanup stage similar to what we propose.  } 
Our proposal is based on the premise---supported by empirical observation---that the substitution-based representation clearly identifies the ``core'' senses that explain most of the mass in the data (as evident by the resulting high task scores), but also introduces some noise around more niche usages or less clear-cut instances.

We thus follow a strategy in which we provide a relaxed \emph{upper-bound} on the number of senses (we use 10), induce this number of clusters, and mark each cluster as being either \emph{weak} or \emph{strong}.  We then discard the weak clusters, merging each of them into a corresponding strong cluster. 

For each target, we induce a soft clustering of the corresponding word occurrences into a fixed number of $c=10$ senses. Each instance (word occurrence) is now probabilisticaly associated with $c$ senses. We say that a sense \emph{dominates} an instance if it is the most probable sense for that instance. We identify senses that dominate less then $m=2$ instances and mark them as \emph{weak} senses. The remaining senses (those that dominate $m$ or more instances) are marked as \emph{strong}. 

Recall that each sense is also a hard clustering over representatives. We associate each sense with the average vector of its representatives (centroid).
For each of the weak senses $w$, we find the closest strong sense $s$ to $w$ according to the cosine distance between their centroids, assign $w$s representatives to $s$ and discard $w$. We then re-do the soft clustering of instances based on the set of strong senses and the representatives within them.
\\[0.5em]
\noindent\textbf{Evaluating the dynamic number of senses}
Unfortunately, this dynamic sense number assignment did not improve AVG WSI scores on the SemEval 2013 dataset.
However, eye-balling the results indicates that the method produces reasonable sense induction solutions.
Digging further, we found out that using the \emph{gold} (oracle) number of senses for each

target also had a very minimal effect on the WSI scores ($\sim0.5$ AVG addition). 
The AVG score in the SemEval 2013 WSI task is the geometric mean of the FNMI metric and the FBC metric, where the first one prefers many small clusters, while the second one prefer fewer and larger clusters. Neither of FNMI, FBC and AVG are sufficient for indicating a good number of clusters.\footnote{The story for the SemEval 2010 WSI task metrics is similar, with V-Measure favoring smaller clusters and F-S favoring larger ones.}

The metrics also do not penalize over-specification of small senses:
while FBC and F-S should discourage over-specification, their measures are proportional in instance pairs and would not punish small mass perturbations, even if those produce an excessive number of senses. We thus aim for a more direct measure for evaluating the produced number of sense-clusters.

Previous work, e.g. \cite{song2016word,hwang2019autosense}, compare the absolute number of senses. 
We instead opt for the somewhat easier task of measuring the \emph{correlation} between the number of induced senses and the gold number of senses. 
 To motivate measuring correlation and not absolute numbers, recall that the SemEval task's
sense-inventory is based on WordNet, whose sense hierarchy is very fine-grained.
For example, it differentiates between \emph{dark} used to describe skin color
and \emph{dark} used to describe objects such as pants. 
A coarser grained WSI solution may constantly produce fewer senses for each target yet still be valuable.\footnote{For example, in the context of query-based search, a user may be satisfied with the coarse grained distinction of \emph{"dark(blackness) times"} and \emph{"dark(sad) times"}. } By measuring \emph{correlation} to the gold number of senses rather than absolute difference we could add invariance to sense granularity.

Our system results in a spearman rank correlation of 0.43$\pm$0.05 (all p-values $<$ 0.03) with the gold number of senses on
SemEval 2013.
Is that a good number? To put the number in context, we compare it to the
correlation of the gold WordNet senses to a human-created coarser sense
inventory: senses in the New Oxford American Dictionary --- NOAD \citep{noad-dict}. We map from
WordNet to NOAD by using the work of \citet{naod} who annotated the SemCor
corpus with NOAD senses in addition to the existing WordNet ones.\footnote{We
take the most probable NOAD sense to each WordNet one, according to the
parallel corpus. This reduces the 399 senses SemEval to 205 (89 of which were
not found in SemCor and left intact). Overall, 87\% of the tokens were mapped to
their coarse grained NOAD senses.}

As expected, comparing the solution obtained by this oracle mapping to the
SemEval 2013 gold labels result in very high WSI scores (FNMI 52.1, FBC 84.7,
AVG 66.4). More interesting is the correlation of sense numbers: 0.47, compared to
0.43 obtained by our method. Our method spearman rank correlation to the number of senses in the \emph{NOAD} labels is 0.44.

\subsection{Cluster Interpretability}
The substitution-based method lends itself to introspection by considering the substitutes. We highlight the most prominent and informative word substitutes for each sense by computing the pointwise mutual
information (PMI) between substitute words and sense clusters. We then annotate each
sense with its top 10 most associated substitutes (its \emph{signature}). These sense signatures can be said to present the essence of what is captured by
each sense cluster.
As an example, one induced sense for the target \emph{meet(VERB)}\footnote{Additional examples are provided in the appendix.} 
is characterized by the words ``convene'', ``group'', ``crowd'', indicating the sense of a meeting that involves many participants. Interestingly, the WordNet \emph{meet(VERB)} entry \emph{does not} make such a distinction between meeting types by the number of their participants, highlighting a case were the unsupervised algorithm refined the human curated lexicon.
Inspecting clusters and their signatures allows us to identify good and bad clusters,

and identify failure modes in its process, as we do in the next section. 

\section{Detailed Error Analysis}
\vspace{-5pt}
\label{sec:analysis}
Armed with the cluster signatures, we turn to manually inspect all the produced sense clusters and their associated words. We  identify the following characteristic failure modes:
\\\textbf{LM}: errors of the underlying BERT LM;\footnote{Often resulting on transcribed speech, a domain BERT was not trained on.}
\\\textbf{SPLIT}: an additional cluster for an existing sense, for example the
sense \emph{encouraging, close, personal, ...} for \emph{warm(ADJ)} when the sense
\emph{compassionate, favorable, kind, ...} already exists;
\\\textbf{TEMPLATE}: substitutes rely excessively on a template-like pattern;
\\\textbf{TOPIC}: substitutes rely excessively on topical words;
\\\textbf{MERGE}: cluster includes several distinct senses;
\\\textbf{OTHER}: cluster includes an incoherent mix of multiple senses with
incoherent substitutes.

\begin{figure}[t!]
\centering
      \includegraphics[width=0.8\linewidth]{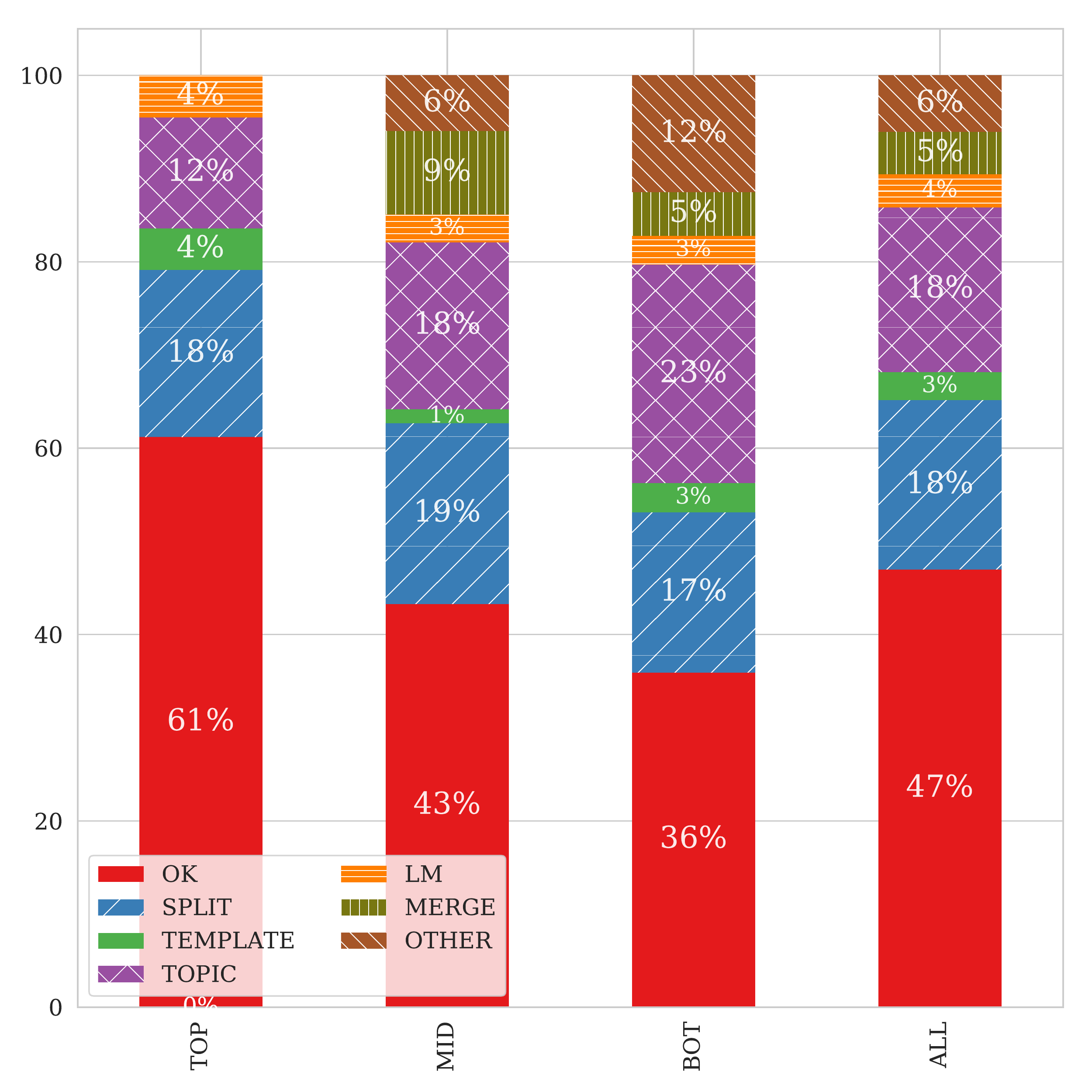}
    \caption{Manual error analysis. Percent of error cases broken down by task score. TOP are best 20\% preforming targets, MID and BOT are third and last 20\% splits accordingly. }
    \label{fig:qualbd}
\end{figure}

We sort the SemEval 2013 targets according to our accuracy on them, and consider
the 20\% top scoring targets (TOP), 20\% middle scoring (MID), and 20\% bottom scoring (BOT), each containing 10 targets.
For each of these groups, we inspect all induced senses, and manually categorizing to
the above failure cases, or to OK in case they are correct. Figure \ref{fig:qualbd}
summarizes the results.\footnote{The supplementary material shows examples of the analyzed cases as well as suggestions for handling the identified failure cases in future work.} A clear trend is that the majority of issues relate to splitting and merging of clusters and relying on topical substitutions, while language-modeling and template-following problems are far less severe.

\section{Conclusion}
\vspace{-0.4em}
We improved a recent WSI method by allowing it to produce a dynamic number of senses, and by showing how the resulting clusters could be inspected and validated through the identification of per-sense characteristic substitutes. These are then used to perform error analysis of the method and its culprits, highlighting the major modes of failure and their prevalence, suggesting promising avenues for future work.
Additionally, incorporating BERT as an LM improves the state of the art in two recent SemEval WSI tasks by a large margin, and    validates the utility of the dynamic-patterns approach of \citet{amrami2018word}.

\bibliography{main}
\bibliographystyle{main}

\appendix

    \newpage
     \section{Handling of identified failure modes} 
Some of the failure modes mentioned in section 4 can be remedied by various means.

LM and TEMPLATE cases are relatively rare and stand out when debugging the final solution. Their distinct distribution usually pushes them into clusters of their own, allowing identification and possibly their removal before rerunning the procedure.

Using a suitable LM for the target domain is important, and indeed most LM
failure we encountered are due to transcribed spoken text. Fine-tuning BERT on the
domain of interest could improve results.

The MERGE and TOPIC cases deal with the discerning resolution of our method. 
An interesting direction for future work is finding a way to collect additional target usages to better model the borderline cases. This also seems like a promising direction to take with SPLIT cases. 

The OTHER classified senses are cases where our method completely fails. These
include targets such as \emph{become(VERB)} which are
indeed hard to sense-induce without some mental process, specifically with substitutions alone. For example, the WordNet senses differentiate between ``become: a change in state'' and ``become: transform into something else'', similarly to Spanish's ser/estar distinction. 


\section{Quality analysis examples} 
\label{app:examples}
The following provides examples of the different error categories we use, as well as demonstrates the senses that are induced by the method for some cases, and their descriptions according to the PMI method.  

Each table shows an induced sense, its high-ranking PMI words, and sentences associated to this sense.
We additionally provide our assessment of that sense (OK, SPLIT, MERGE, TOPIC, TEMPLATE, LM, OTHER), as well as the gold-label WordNet sense for each sentence.

We begin with the senses for meet(VERB), a target that our method scores high on (Table \ref{tab:examples}).  
We follow with wait(VERB), a target which our methods scores low on (Table \ref{tab:examples2}).
Finally, Table \ref{tab:qualrest} demonstrates the error categories not present in the previous ones.  

\begin{table*}[htp!]
\centering
\begin{tabularx}{\linewidth}{| c |  c| X | }
	\hline
	\textbf{No.} & \textbf{Class} \ & \textbf{High PMI words} \\ \hline \hline 
    1 & OK &  hug, pick, thank, ., ..., surprise, impressed, marrie, hire, welcome \\ \hline
    \multicolumn{3}{|p{0.92\linewidth}|}{
    \begin{itemize}[topsep=0pt,itemsep=0pt,partopsep=0pt, parsep=0pt, rightmargin=0pt]
        \item You're going to {\bf meet} John Speckman! (5)
        \item So I guess one question might be how I {\bf met}
  my wife. (1)
      \item We are taking you to {\bf meet} him the day after you arrive. (1)
    \end{itemize}
  } \\
  \hline\hline
  2  & OK & qualify, offer, below, violate, comply, accomplish, supply,complete,fill,accommodate\\ \hline
    \multicolumn{3}{|p{0.92\linewidth}|}{
    \begin{itemize}[topsep=0pt,itemsep=0pt, parsep=0pt]
        \item And we need Your help to {\bf meet} the challenge! (3)
        \item So, I want to thank you on {\bf meeting} my first condition. (3)
      \item They could not {\bf meet} conditions if their competitors were free to ignore them. (3)
    \end{itemize}
  }\\
  \hline\hline
    3 & SPLIT & conversation, summit, friendly, discussion, business, spend, touching, partner, dining \\ \hline
    \multicolumn{3}{|p{0.92\linewidth}|}{
    \begin{itemize}[topsep=0pt,itemsep=0pt, parsep=0pt]
        \item it's gonna make the people they're {\bf meeting} with feel very uncomfortable ... (5)
        \item Best wishes until we {\bf meet} again-perhaps over Volume 9 ... (5)
      \item He and Atta agreed to {\bf meet} later at a location to be determined. (7)
    \end{itemize}
  }\\
  \hline\hline
    4  & OK & group, convention, weekly, schedule, parliament, convene, celebrate,crowd,originate \\ \hline
    \multicolumn{3}{|p{0.92\linewidth}|}{
    \begin{itemize}[topsep=0pt,itemsep=0pt, parsep=0pt]
        \item A group called the League of Prizren, named for the Kosovo town where it {\bf met}, ... (7)
        \item cat and bagpipean society a society which {\bf met} at their office ... (7)
      \item A summer Antiekmarkt or antique market {\bf meets} at Nieuwmarkt on Sundays ...(7)
    \end{itemize}
  }\\
  \hline\hline
    5 & OK & direct, encounter, dare, oppose, reaction, repulse, cause, underwent, face, react \\ \hline
    \multicolumn{3}{|p{0.92\linewidth}|}{
    \begin{itemize}[topsep=0pt,itemsep=0pt, parsep=0pt]
        \item They were greeted as liberators by the peasants and {\bf met} only desultory resistance ... (4,9)
        \item ... astounded by the funny logic of, say, {\bf meeting} one's match ... (9)
      \item It's too bad that ... this understanding has to {\bf meet} with such hostility, don't you think? (4)
    \end{itemize}
  }\\
  \hline\hline
    6 & OK & maximum, phase, interval, curve, origin, converge, respectively, cancel, border, dip \\ \hline
    \multicolumn{3}{|p{0.95\linewidth}|}{
    \begin{itemize}[topsep=0pt,itemsep=0pt, parsep=0pt]
        \item we can draw a line of those tangencies ..., that  {\bf meet} at the initial apple-pear distributions... (6)
    \end{itemize}
  }\\
  \hline\hline
  7& TOPIC & investigate, cost, fund, ease, budget, recover, shoulder, offset, slash, decrease \\ \hline
    \multicolumn{3}{|p{0.92\linewidth}|}{
    \begin{itemize}[topsep=0pt,itemsep=0pt, parsep=0pt]
        \item  ... appealed to the state government to help {\bf meet} the cost of burying armed robbers...(3)
    \end{itemize}}\\
  \hline\hline
    \multicolumn{3}{|p{0.95\linewidth}|}{
    WordNet senses for meet(VERB) in gold labels:
    \begin{multicols}{2}
    \begin{enumerate}[topsep=2pt,itemsep=0pt,partopsep=0pt, parsep=0pt]
        \item get to know; get acquainted with 
        \item collect in one place
        \item fill or meet a want or need
        \item experience as a reaction
        \item come together
        \item be adjacent or come together
        \item get together socially or for a specific purpose
        \item meet by design; be present at the arrival of
        \item contend against an opponent in a sport, game, or battle
        
    \end{enumerate}
    \end{multicols}}
    
    \\
  
  \hline
\end{tabularx}
\caption{Senses induced for the target meet(VERB) on which our method perform relatively well. At the end of each sentence in parentheses is its tagged Word Net sense in the gold labels. In our manual inspection, Sense \#1 seems semantically close enough to be classified OK. Sense \#3 is classified as SPLIT due its similarity to sense \#1. Sense \#7 is classified as TOPIC due to its substitutes which are not lexical. It's interesting to see the method induced a sense for group meeting, sense \#4, with substitutes such as convene and crowd. }
  \label{tab:examples}
\end{table*}

\begin{table*}[ht!]
\centering
\begin{tabularx}{\linewidth}{| c | c | X  |}
	\hline

	\textbf{No.} & Class & \textbf{High PMI words} \\ \hline \hline 
    1 & OK & lodge, cool, bray, rock, glide, bend, hush, groom, camp, creep \\ \hline
    \multicolumn{3}{|p{0.92\linewidth}|}{
    \begin{itemize}[topsep=0pt,itemsep=0pt,partopsep=0pt, parsep=0pt]
        \item The horses {\bf wait} under the cooling shade for their next customers. (1)
    \end{itemize}
  } \\
  \hline\hline
  2 & $\sim$OK & bench, staff, guest, bounce, pat, pit, ticket, fare, to, other \\ \hline
    \multicolumn{3}{|p{0.92\linewidth}|}{
    \begin{itemize}[topsep=0pt,itemsep=0pt,partopsep=0pt, parsep=0pt]
        \item ... offer the best sightlines, roomier seats, and {\bf wait} staff who peddle gourmet fare. (3)
    \end{itemize}
  }\\
  \hline\hline
    3 & OTHER & reasonable, slack, qualify, delivery, short, a, temporary, week, hesitation, due \\ \hline
    \multicolumn{3}{|p{0.92\linewidth}|}{
    \begin{itemize}[topsep=2pt,itemsep=0pt,partopsep=0pt, parsep=0pt]
        \item ... without the need to {\bf wait} until everyone is in town for a meeting. (4)
        \item ... operator will be paid at some average earnings rate during the {\bf waiting} period. (4)
      \item ... and uh i would agree a a short {\bf waiting} period would be appropriate to uh ... (4)
    \end{itemize}
  }\\
  \hline\hline
    4  & TOPIC & literally, cooking, everything, pregnant, family, lot, town, money, forever, food \\ \hline
    \multicolumn{3}{|p{0.92\linewidth}|}{
    \begin{itemize}[topsep=0pt,itemsep=0pt,partopsep=0pt, parsep=0pt]
        \item He had a farm {\bf waiting} for him right? (2)
        \item If Clinton, ... , was a time bomb {\bf waiting} to explode, then ... (1)
      \item ... as you wouldn't if you had a wife who looked like that {\bf waiting} for you. (1)
    \end{itemize}
  }\\
  \hline\hline
    5  & LM & ago, they, fade, along, since, drank, though, afterwards, sometimes, uh \\ \hline
    \multicolumn{3}{|p{0.92\linewidth}|}{
    \begin{itemize}[topsep=0pt,itemsep=0pt,partopsep=0pt, parsep=0pt]
        \item ... i sang in a couple of uh community choirs and then um {\bf waited} for a while ...(4)
    \end{itemize}
  }\\
  \hline\hline
  6  & OTHER& write, reach, argue, appear, bother, act, seem, star, [, wish \\ \hline
    \multicolumn{3}{|p{0.92\linewidth}|}{
    \begin{itemize}[topsep=0pt,itemsep=0pt,partopsep=0pt, parsep=0pt]
        \item  I'm not sure why you {\bf waited} a week.(4)
        \item  Good things come to those who {\bf wait}.(4)
        \item  He had been {\bf waiting} for Oedipa in the bathroom. (1)
    \end{itemize}}\\
  \hline\hline
    \multicolumn{3}{|p{0.92\linewidth}|}{
    WordNet senses for wait(VERB) in gold labels:
    \begin{enumerate}
        [topsep=2pt,itemsep=0pt,partopsep=0pt, parsep=0pt]
        \item stay in one place and anticipate or expect something
        \item look forward to the probable occurrence of
        \item serve as a waiter or waitress in a restaurant
        \item wait before acting
    \end{enumerate}}\\
  
  \hline
\end{tabularx}
\caption{Senses induced for the target wait(VERB) on which our method perform poorly. In our manual inspection, sense \#6 is classified as OTHER, this class of found senses usually group up a large portion of unrelated sentence, making their differentiating substitutes an incoherent bag of left-overs. Sense \#2's substitutes aren't very informative for its sense ("serve as a waiter") but they do distinguish this sentence from the other sensed sentences.}
  \label{tab:examples2}
\end{table*}

\begin{table*}[ht!]
\centering
\begin{tabularx}{\linewidth}{| c | c | X  |}
	\hline

	\textbf{Target} & \textbf{Class} & \textbf{High PMI words} \\ \hline \hline 
    strike(VERB) & MERGE & rally, roar, mobilize, go, picket, dominate, rise, uprising, rebel, riot \\ \hline
    \multicolumn{3}{|p{0.9\linewidth}|}{
    \begin{itemize}[topsep=0pt,itemsep=0pt,partopsep=0pt, parsep=0pt]
        \item William Safire Language Maven {\bf strikes} Again.
        \item ...  in 1953 when the workers {\bf struck} in Berlin and the Party told them to stop ...
        \item ...   on Feb. 21, 1868, the Radicals {\bf struck}. ...
    \end{itemize}
  } \\
  \multicolumn{3}{|p{0.9\linewidth}|}{
  This sense cluster the meaning of two unrelated concepts: "strikes again" as in "does it again", and "strike" as in "worker uprising". The characteristic substitutes are still informative.
  }\\
  \hline\hline
  sight(NOUN) & TEMPLATE &  advance, propose, glamour, diamond, lipstick, wing, aspiration \\ \hline
    \multicolumn{3}{|p{0.9\linewidth}|}{
    \begin{itemize}[topsep=0pt,itemsep=0pt,partopsep=0pt, parsep=0pt]
        \item ... Barbara Hershey, 50, set her {\bf sights} on Brooks' fiance ...
        \item The Enquirer provides details about how Monica set her {\bf sights} on her man ...
        \item ... Frank Sinatra's widow, Barbara, has set her {\bf sights} on Grace Kelly's widower, ...
    \end{itemize}
  }\\
  \multicolumn{3}{|p{0.9\linewidth}|}{
  This template of X set her {\bf sights} on Y, produces a distinct substitutes distribution which pushes the sentences above into a sense of their own. This "over-fits" the target usage, giving weight to substitutes such as "lipstick" in the example above.
  }\\
  \hline
\end{tabularx}
\caption{Examples of error cases not appearing the previous two clustering solutions.}
  \label{tab:qualrest}
\end{table*}

\end{document}